\documentclass[letterpaper, 10 pt, conference]{ieeeconf}  

\IEEEoverridecommandlockouts                              
                                                          
\usepackage[greek,english]{babel}
\usepackage{dirtytalk}
\usepackage{cite}
\usepackage{amsmath,amssymb,amsfonts}
\usepackage{algorithmic}
\usepackage{graphicx}
\usepackage[utf8]{inputenc}
\usepackage{csquotes}
\usepackage{textcomp}
\usepackage{booktabs}
\usepackage[capitalize]{cleveref}
\usepackage{diagbox}
\usepackage{array}
\usepackage[table,xcdraw]{xcolor}
\usepackage{threeparttable} 
\usepackage{multirow}
\usepackage{lscape}
\usepackage{xcolor}
\usepackage{float} 

\title{\LARGE \bf
A Taxonomy of Semantic Information in Robot-Assisted Disaster Response
}

\author{Tianshu Ruan$^{1*}$ Hao Wang$^{1}$ Rustam Stolkin$^{1}$ and Manolis Chiou$^{1}$
\thanks{This work was supported by EPSRC grants EP/R02572X/1, EP/P01366X/1, and EP/P017487/1.}
\thanks{$^{1}$Extreme Robotics Lab (ERL), University of Birmingham, UK}
\thanks{$^{*}$\tt\small txr094@student.bham.ac.uk}
}

\begin{document}

\maketitle
\thispagestyle{empty}
\pagestyle{empty}

\begin{abstract}

This paper proposes a taxonomy of semantic information in robot-assisted disaster response. Robots are increasingly being used in hazardous environment industries and emergency response teams to perform various tasks. Operational decision-making in such applications requires a complex semantic understanding of environments that are remote from the human operator. Low-level sensory data from the robot is transformed into perception and informative cognition. Currently, such cognition is predominantly performed by a human expert, who monitors remote sensor data such as robot video feeds. This engenders a need for AI-generated semantic understanding capabilities on the robot itself. Current work on semantics and AI lies towards the relatively academic end of the research spectrum, hence relatively removed from the practical realities of first responder teams. We aim for this paper to be a step towards bridging this divide. We first review common robot tasks in disaster response and the types of information such robots must collect. We then organize the types of semantic features and understanding that may be useful in disaster operations into a taxonomy of semantic information. We also briefly review the current state-of-the-art semantic understanding techniques. We highlight potential synergies, but we also identify gaps that need to be bridged to apply these ideas. We aim to stimulate the research that is needed to adapt, robustify, and implement state-of-the-art AI semantics methods in the challenging conditions of disasters and first responder scenarios.

\end{abstract}




\section{Introduction}
Detecting and understanding semantic information (i.e. the meaning of things) could enhance robots' abilities to perform complex tasks. Humans can intuitively understand semantics, infer context, and incorporate it into ``situational awareness'' (SA). In contrast, semantic understanding (i.e. the process of understanding semantic information) in robotics remains a challenging and open research problem.  

The term semantics is derived from the Greek verb \textgreek{σημαίνω} [sēmainō] meaning \say{to signify} or \say{to mean}\footnote{https://www.britannica.com/science/semantics/Historical-and-contemporary-theories-of-meaning}. Typically in robotics and Artificial Intelligence (AI), semantics refers to the understanding of things, environments, and situations. Garg et al. define \say{semantics in a robotic context to be about the meaning of things: the meaning of places, objects, other entities occupying the environment, or even the language used in communicating between robots and humans or between robots themselves} \cite{garg2020semantics}. Additionally, in computer vision, the Australian Institute for Machine Learning (AIML) states that \say{semantic vision seeks to understand not only what objects are present in an image but, perhaps even more importantly, the relationship between those objects} \cite{semanticvision}. 

In the context of this paper, we define semantics as: \textit{extracting and conveying an understanding of environments, to both robots and humans, in terms of information about entities and both explicit and implicit relationships between entities}. For utility, our working definition aims to be somewhat explicit and restrictive in terms of understanding of the environment. It comprises two key features of the environment: entities and the relationships between them. Note that this implies two levels of information and abstraction, i.e. relationships are abstract concepts that link two or more physical entities.


In this paper, we focus on semantics in disaster and emergency response, and tasks in hazardous environments. Robots are increasingly being used in hazardous or ``extreme environment'' industries and have become increasingly embedded in emergency response teams. Robots perform a wide range of tasks, including reconnaissance and remote inspection in hard-to-reach or dangerous zones, as well as heavy-duty manipulation or transporting tasks. For safety reasons, humans should be located remotely from hazardous zones, but must rapidly assess the circumstances where a robot is operating. With currently deployed robots being mostly teleoperated and lacking advanced AI capabilities, it is notoriously difficult for human operators to gain SA from limited video and sensor data \cite{manolisKHGiros2022}.

Semantic understanding is an important element in building a clear SA that is crucial for humans, robot autonomy \cite{murphy2005up}, and Human-Robot Interaction (HRI) \cite{smith2013human}. Given onboard semantic understanding capabilities, a remote robot could: i) make better and more complex autonomous decisions; and/or ii) provide enhanced information to assist human operators with clear and rapid SA. Currently, much of semantics-related AI research is focused on semantic image understanding \cite{garg2020semantics}, typically using deep neural networks, such as ``semantic segmentation'' \cite{FCN, Ronneberger2015UNetCN}, or object detection and recognition \cite{girshick2015fast, wang2022yolov7}. Related research also includes semantic Simultaneous Localization and Mapping (SLAM) \cite{rosinol2021kimera} and 3D reconstruction of a scene while identifying and labeling types of materials \cite{zhao2017fully}.


However, despite advances in disaster response robotics, and parallel growth in AI research on semantic understanding, there has been comparatively little transfer from one domain to the other. Partly this is because rapid AI advances in semantics are recent developments and the technology remains predominantly ``low-TRL'' (Technology Readiness Level). In contrast, disaster environments are profoundly unstructured and difficult. ``High-consequence'' rescue tasks demand extremely robust, high-TRL equipment \cite{manolisKHGiros2022}. Additionally, applications of e.g. semantic segmentation in computer vision have tended to focus on domains such as driverless cars, which offer a large market and capture more attention than the specialized first responder domain.
 
This paper offers a step towards linking AI semantics research with disaster robotics R\&D. Contributions include: i) identifying end-user semantic information needs and potential synergies; ii) identifying semantics research gaps to be bridged; iii) framing a coherent structure for thinking about these issues, in terms of a taxonomy that connects the work of these two research communities. Specifically, we address the following three research questions:
\begin{itemize}
\item\textbf{RQ1)} What semantic information is useful in the robotic disaster response operation?

\item\textbf{RQ2)} How semantic information is structured in the robotic disaster response domain?

\item\textbf{RQ3)} What AI-extracted semantic information can be feasibly used in disaster robotics field deployments and how can we obtain it?
\end{itemize}

\section{Common Tasks for Robots in Disaster Response} \label{Tasks}

We first survey robot deployments in the response and recovery phases \cite{murphy2014disaster, schneider2015possible} of disasters from the perspective of commonly executed tasks. A variety of different robotic platforms have been deployed after disasters, e.g. floods, earthquakes, hurricanes, mudslides, or nuclear incidents. This discussion here is not meant to be exhaustive, but rather indicative, to facilitate identifying useful semantics linked to required robot tasks. 

Unmanned Aerial Vehicles (UAVs) can perform tasks that include mapping floods, additional flood risk estimation \cite{murphy2016two}, damage, or ground search \cite{fernandes2019quantitative}. Furthermore, UAVs can offer high-resolution remote imagery when satellites or manned aircraft are unavailable \cite{fernandes2018quantitative} and help inspect dams, bridges, buildings, or monitor the risk potential of consecutive disasters \cite{murphy2016use}.

Unmanned Ground Vehicles (UGVs) are often used to gain access to environments that are dangerous for humans to reach \cite{casper2003human}. UGVs can be assigned to build 3D maps of the interior of buildings \cite{kruijff2012rescue}, assess damage to buildings \cite{kruijff2016deployment}, monitor temperature \cite{nagatani2013emergency}, radiation levels \cite{ConnorAdata}, retrieve samples (e.g. contamination) \cite{manolisKHGiros2022}, or search for victims \cite{micire2008evolution}. In addition to reconnaissance tasks, UGVs are deployed to install devices, serve as communications relay stations, or provide camera views to control other robots \cite{manolisKHGiros2022}. In addition, heavy-duty UGVs can be used for cutting or moving structures and debris, or, e.g., removing inflammable gas cylinders from industrial buildings during fires.

Unmanned Surface or Underwater Vehicles (USVs and UUVs) are deployed after floods and hurricanes \cite{murphy2014disaster}, oil spills, or shipwrecks. They can deploy cameras and sonars to detect damage to structures, locate submerged debris \cite{micire2008evolution}, estimate the volume of water in rivers and streams \cite{fernandes2018quantitative}, and deliver supplies \cite{scerri2011flood}. 
\begin{figure}[H]
    \centering
    \includegraphics[scale =0.35]{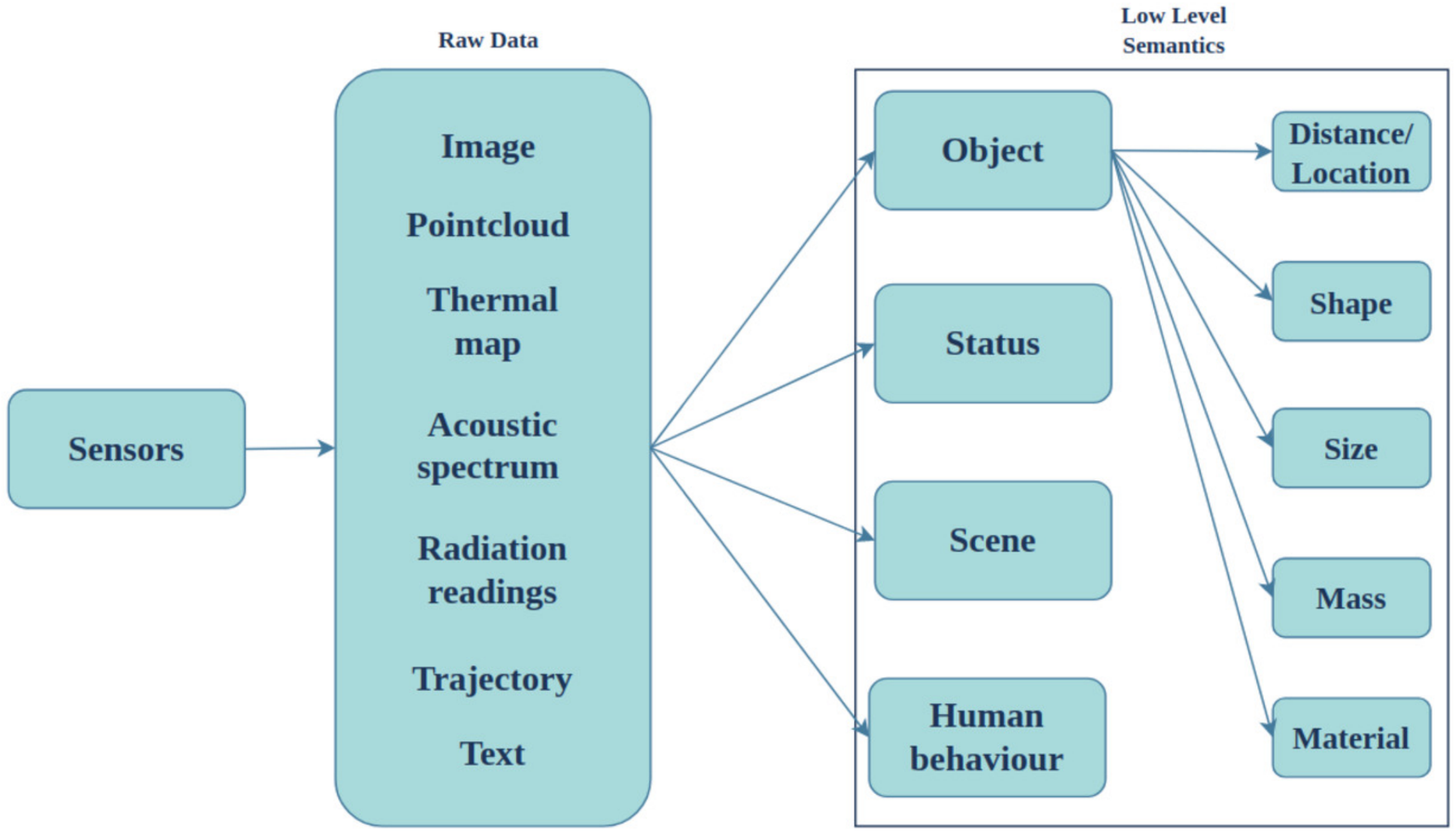}
    \caption{The relationship between the low-level semantics and raw data.}
    \label{fig: Raw data and low-level semantics}
\end{figure}

We summarize key tasks in response and recovery phases into four main categorizes:

\begin{itemize}

\item \textbf{Exploration and reconnaissance (ER) tasks} such as mapping, hazardous materiel (HAZMAT) detection, sampling, and structural inspection \cite{kruijff2012rescue,micire2008evolution,fernandes2018quantitative,murphy2016use, manolisKHGiros2022}.

\item \textbf{Auxiliary tasks} such as acting as a mobile beacon or communications relay/repeater \cite{agha2021nebula,manolisKHGiros2022} 

\item \textbf{First-aid tasks} such as in situ medical assessment and medically sensitive extrication \cite{murphy2014disaster}.

\item \textbf{Heavy-duty tasks} such as rubble removal, shoring unstable rubble, providing logistic support, fire extinguishing, switching the valves and breaching obstacles \cite{murphy2004trial,nagatani2013emergency}.
\end{itemize}

\section{Semantic Information Taxonomy} \label{semantics}

Here we present our proposed taxonomy which is based on the semantic information needed in the common tasks (See \cref{Tasks}). We group the corresponding semantics into two main categories from the perspective of the human's understanding of different abstractions: low-level semantics and high-level semantics. Low-level semantics refers to information that is obvious and readily available with less effort to humans (e.g. object detection). High-level semantics refers to an abstract understanding of the environment that humans need further evaluation to infer (e.g. risk of hazardous situations). Lastly, we discuss the relationship between low-level semantics, high-level semantics, context, and the methods to obtain them.


\subsection{Low-level semantics} \label{low}
Low-level semantics reveals the most salient features of the environment. Salient features mean that they can be noticed and understood by human's direct intuition without further thinking, rather than mental effort for inference. We categorize low-level semantics into the following categories:

\begin{itemize} 

\item \textbf{Objects:} detection of HAZMAT, victims, and other significant features of the environment \cite{sharifi2020deephazmat}. Some detailed sub-category low-level semantics can also be used to describe the objects (e.g. distance and material).

\item \textbf{Status:} a more detailed condition of the detected objects, e.g., cracks or delamination of concrete \cite{pozzer2021semantic}, or object on fire.

\item \textbf{Scene understanding:} an identification of the types of room or space which is significant for high-level semantics processing, e.g. if the space is an office or warehouse \cite{shi2021semantic}.

\item \textbf{Human behaviour:} detection of human's movements in the post-disaster scenario, e.g. dangerous movements of victims \cite{jang2020detection}.

\end{itemize} 

All the above categories help robots and humans to build an accurate SA and support tasks including ER tasks, first aid tasks and heavy-duty tasks (see \cref{Tasks}).

We summarize the low-level semantics in \cref{fig: Raw data and low-level semantics}. Note that a single low-level semantic feature can be potentially processed from multiple raw data. We can obtain the same low-level semantics from different sensors. For instance, we can obtain scenes from either cameras, microphones, or LiDar. Additionally, there is detailed sub-category low-level semantics (e.g. distance, material, and mass) constituting the understanding of objects. 
The fusion among diverse types of raw data can extend the types of low-level semantics. 

\subsection{High-level semantics} \label{high}

\begin{figure}[H]
    \centering
    \includegraphics[scale =0.35]{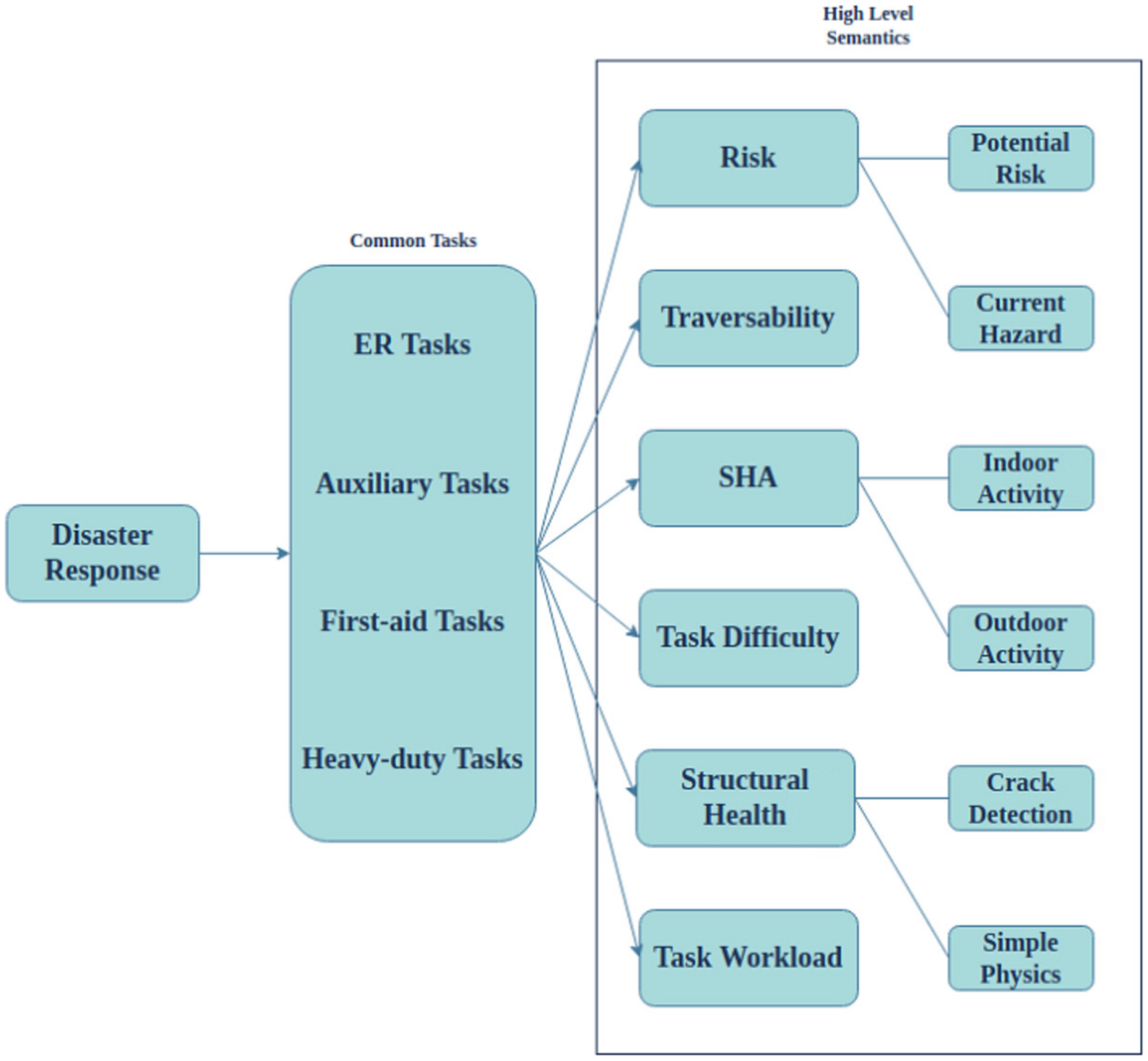}
    \caption{The proposed high-level semantics in relation to the common disaster robotic tasks.}
    \label{fig: Tasks and high-level semantics}
\end{figure}

High-level semantics is more complex than low-level semantics. It usually comes from the evaluation, estimation or inference of a human's perspective about the onsite situation for robots and humans, rather than direct intuition. We group high-level semantics into the following main categories that are useful for disaster robotic operations (See \cref{fig: Tasks and high-level semantics}): 

\begin{itemize} 

\item \textbf{Risk:} an assessment of the existing and/or the potential danger of the environment to the humans and the robots. 
\item \textbf{Traversability:} the difficulty for humans or robots to travel through an area.
\item \textbf{Signs of human activity (SHA):} the inference or estimation that victims or survivors can be potentially found in the scene.
\item \textbf{Task difficulty:} the prediction of the difficulty for robots or humans to complete the given task.
\item \textbf{Structural health:} the indication or assessment of how stable a structure is, e.g. the structure of debris or a building. 
\item \textbf{Task workload:} the estimation of how much work is required by the robots and first responders to complete the scheduled tasks.

\end{itemize} 


We further divide risks, SHA, and structural health into more detailed subcategories. The category risk consists of potential risks and current hazards. The SHA is divided into indoor and outdoor activities because of the huge environmental differences. Indoor human activities can be inferred from personal belonging detection and indoor scene understanding, while outdoor human activities can be inferred from human trajectories and history population heatmap, e.g. from prior data in the cloud. Structural health can be determined by crack detection and "simple physics" which refers to the inference to supporters of a heap of objects only based on vision. Task difficulty is divided in terms of semantics fusion e.g. the extent of task completion, traversability of environment, expected time cost, and workload.


High-level semantics requires information from low-level semantics and raw data as well. For instance, risks and SHA can be affected by detected objects, distance readings, and scenes. Moreover, high-level semantics might need additional information from other high-level semantics. For instance, task difficulty due to its complexity might not be accurately estimated just by low-level semantics. Other high-level semantics (e.g. workload and traversability) can also be considered as factors. Hence, high-level semantics requires information from multiple inputs including raw data, low-level semantics, and other high-level semantics.

\subsection{Relationship among low-level semantics, high-level semantics, and context}

Low-level semantics is processed from raw data. In contrast, high-level semantics is processed from the fusion of different low-level semantics, additional high-level semantics, and raw data. High-level semantics processing relies on more different inputs than low-level semantics processing. Hence, low-level semantics is one of the bases of high-level semantics processing. 



\begin{figure}[H]
    \centering
    \includegraphics[scale =0.35]{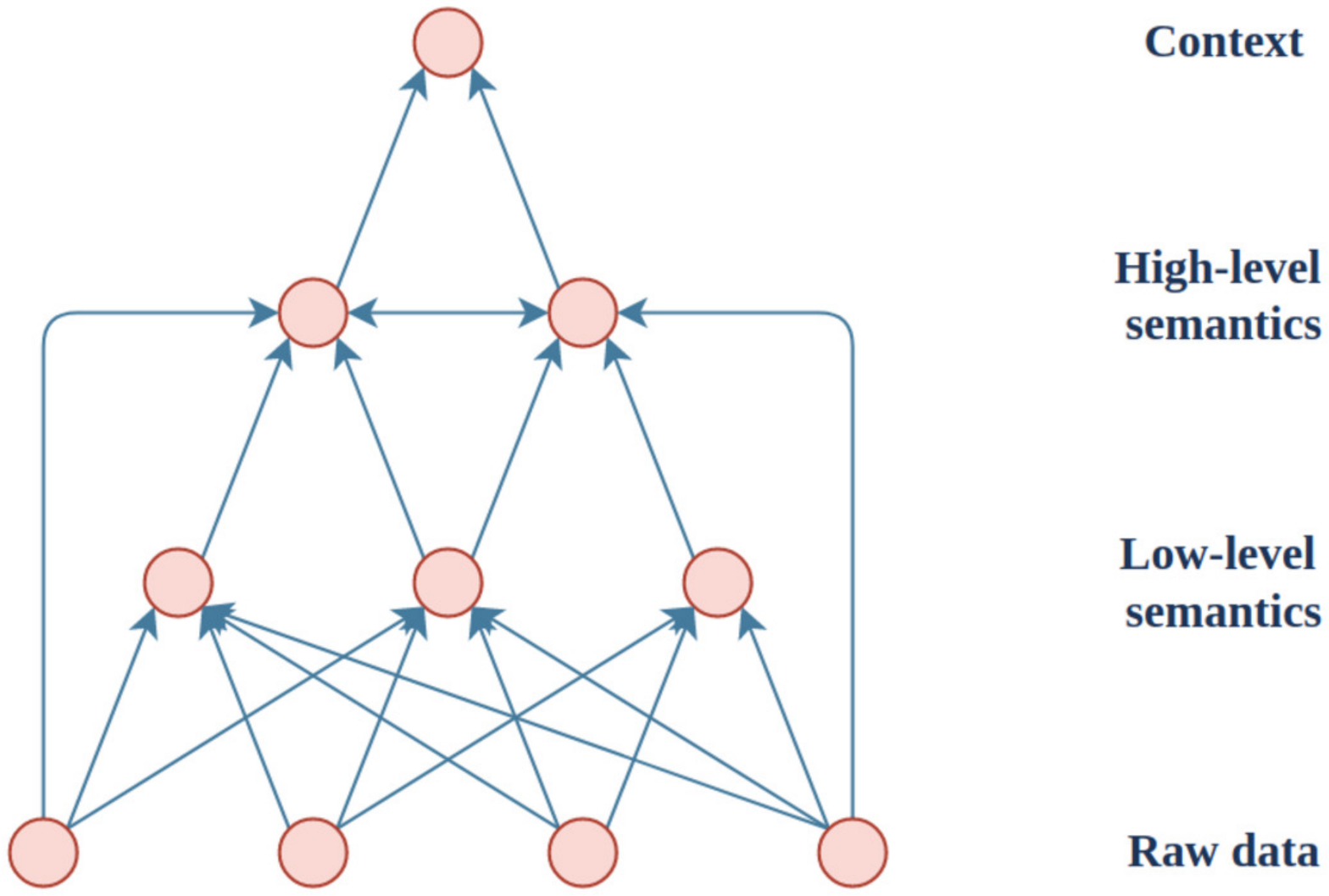}
    \caption{Different levels of information abstraction: each level of information is supported by the lower levels of information.}
    \label{fig: Context}
\end{figure}

Context is often confused with semantics as they have closely related meanings. Context is explained as "context now most commonly refers to the environment or setting in which something (whether words or events) exists" in Marriam-Webster dictionary \footnote{https://www.merriam-webster.com/dictionary/context}. From our perspective, context is the top level of abstraction, i.e., a larger concept that covers all the contents from low-level semantics to high-level semantics, from historical data to up-to-date information (see \cref{fig: Context}). The context describes a scenario or situation by using its components and the connections among the different components. It emphasizes the relationship among all the components which may be semantic information, and sometimes historical information about the environment. In contrast, semantics denotes only part of the information that constructs the global understanding of the environment.

\section{From the Taxonomy to Implementation}
\begin{figure}[htp]
    \centering
    \includegraphics[scale =0.35]{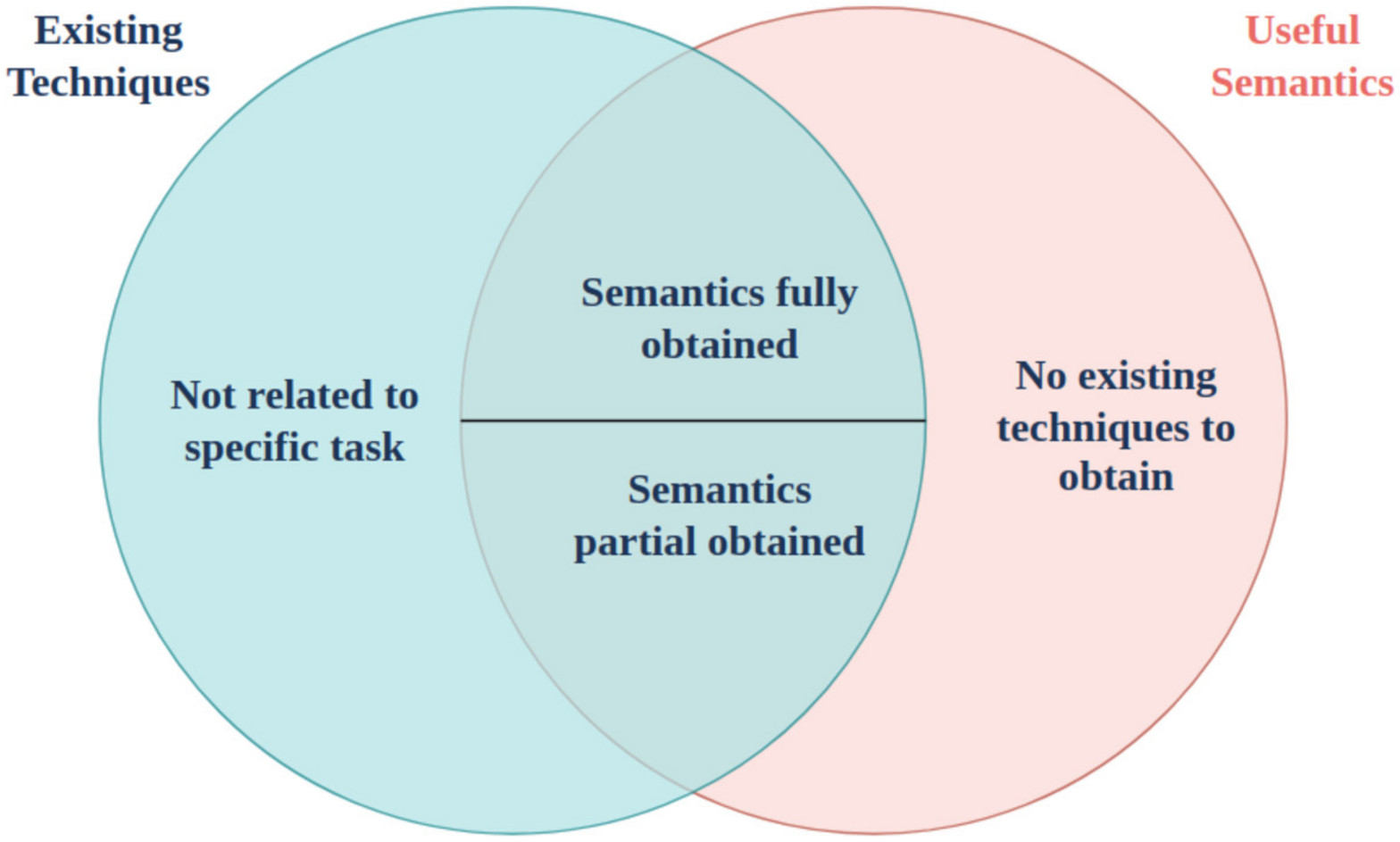}
    \caption{Two aspects of RQ3: the blue circle refers to existing techniques (See Insight I) and the red circle refers to useful semantics (See Insight II). The covered patch illustrates the current semantics understanding technique developments on useful disaster response semantics.}
    \label{fig: Two gaps}
\end{figure}

In \cref{semantics}, we introduce the features of semantics and construct the taxonomy about useful semantics in the disaster response. Here we aim to answer RQ3 from two aspects: state-of-the-art semantic understanding techniques deployments in robotic disaster response and useful semantic information implementation in common tasks. We investigate the question in \cref{fig: Two gaps} and obtain two corresponding insights:

\begin{itemize} 

\item \textbf{Insight I}: The AI community has developed large numbers of semantic understanding techniques. However, few of them have been deployed in the robotic disaster response operations.

\item \textbf{Insight II}: We identified many useful semantics in disaster response based on the taxonomy, but there is a limited number of existing techniques to obtain these semantics.


\end{itemize}



\subsection{Insight I} \label{low processing and high processing}

Here we indicatively investigate the existing semantic understanding techniques from the perspective of disaster robot deployment.

\subsubsection{Low-level semantic processing}
We look into the low-level semantic processing based on the sensors that are popularly used such as RGB cameras, LiDARs, microphones, thermal cameras, and radiation detectors. 

RGB cameras provide colored images and RGB-D cameras provide in addition, depth data. AI community applies object detection \cite{girshick2015fast} and semantic segmentation \cite{wang2022yolov7} techniques to images to categorize and recognize objects. Besides, images provide more semantics e.g. human body keypoints \cite{bourdev2009poselets}. 

LiDARs are widely used in SLAM and other robotic domains. They provide accurate 2D or 3D point cloud maps. In \cite{peng2022mass}, authors use an attention mechanism (a neural network framework \cite{vaswani2017attention}) to conduct a 3D Lidar pointcloud semantic segmentation. Additionally, some researchers are focusing on LiDAR-only solutions without the help of RGB images \cite{milioto2019rangenet++}.   

Acoustic semantic understanding using microphones has two domains. The first domain is Nature Languages Processing (NLP) which figures out the context and meaning of a dialogue or a conversation \cite{devlin2018bert}. Another domain processes sound and noise from nature. It usually uses spectrogram representation to find embedded semantics like signal of lives underwater \cite{stolkin2007passive}, scene \cite{battaglino2016acoustic}, and location of sound \cite{okutani2012outdoor}. 

Thermal cameras outperform RGB cameras in poor illumination environments. Thermal maps improve the performance of image semantic processing in dark environments where RGB cameras hardly capture clear images \cite{sun2020fuseseg}. In addition to that, people can be easily identified in a complex environment \cite{jeon2015human} and also damage to concrete structures can be detected \cite{pozzer2021semantic}. 

Current research on radiation semantics is limited. Because the radiation readings can not provide further information, if they are not fused with other information. In \cite{bird2018using}, authors use a UGV to generate a geometric and radiometric map of a radioactive facility with the assistance of a LiDAR sensor. It has exciting potential to process semantic information e.g. identifying radiation sources, drawing heat maps, and evaluating dangerous areas.  

\subsubsection{High-level semantic processing}
Since most semantic understanding techniques are focusing on low-level semantic processing, there are only two approaches to obtaining high-level semantics based on current techniques, which are traditional machine learning methods and deep learning methods. 

The traditional machine learning methods are mathematical model-based and typically have two steps. First, building a mathematical model. We can quantify high-level semantics by using low-level semantics, additional high-level semantics, or raw data to build metrics. Second, applying traditional machine learning techniques to identify or classify different high-level semantics. To get different extents of one specific high-level semantics, we can apply rules using a mathematical toolbox, e.g. fuzzy logic, based on human cognition, evaluation, and inference to the designed metrics.



The second approach applies deep learning techniques. It is possible to build an end-to-end deep learning network with all the required input and high-level semantics output. However, the deep learning approach has two major defects: First, although a disaster semantic dataset from remote sensing imagery exists \cite{leach2021data}, we still lack specific semantic datasets supported by expert knowledge from first responders. We do not have any disaster semantic dataset based on the first person view (FPV) of UGV. Moreover, the creation of a new dataset is difficult because it requires expert knowledge and a large amount of handcraft work. Second, deep learning networks increase the computation cost when compared with traditional approaches. This might lead to the degradation of real-time performance on robots with limited computational power deployed in the field.

We summarize state-of-the-art semantic understanding techniques in \cref{Summary of Advanced Semantic Processing Techniques} and we contribute the following findings:  

\begin{itemize}

\item RGB cameras and LiDAR are the most widely used sensors. They are good options to fuse.

\item Most semantic understanding techniques concentrate on object detection and scene understanding. This leads them to be very useful in ER tasks, and heavy-duty tasks, but limited use in other tasks.


\item Real-time performance is not considered by some of the researchers. However, for a disaster robotic operation, real-time performance is quite significant and needs more attention. Some of the computationally cheap techniques can be developed into real-time versions that are compatible with disaster robotics requirements.

\item Due to the limitation of RGB cameras in dusty and poor lighting environments, many researchers focus on multiple sensor fusion to overcome the disadvantages of single-sensor solutions. However, these solutions stay on the laboratory level. Few solutions are applied and they are limited in mapping the environments.




\end{itemize} 

\subsection{Insight II}
The deployment of useful semantic understanding techniques in disaster response operations is still an open problem. Thus, we elaborate on Insight II by comparing both common tasks in the disaster operation and what the current semantic understanding techniques can achieve in \cref{Gaps between task requirements and semantic understanding techniques}, to indicate the open opportunities for deploying semantics in disaster response operations. 

Based on the robot field deployments literature after disasters, a few semantic-related techniques are deployed. That is to say, the application of semantic information is limited. Hence, we define four levels to reveal the semantics applications status: technique exists, not applied (T), partial techniques achieved (P), techniques not exist (N), and not related (-). We provide some references and make examples to illustrate the labels. Roof, ventilator, ac unit, and other building structures or components can be detected \cite{leach2021data}. They are useful in reconnaissance. Hence, we label the object "technique exist, not applied (T)". People are now able to detect the health of concrete via thermal and optical images \cite{pozzer2021semantic}. However, the real-time performance is not verified. Therefore, we considered it a "partial technique achieved (P)". For SHA, even though people can obtain human belongings semantics from the environment, we do not have any solutions to evaluate the intensity of human activity. In this case, we label it "N". The \cref{Gaps between task requirements and semantic understanding techniques} reflects the following: 
\begin{landscape}
    \begin{table}
    \normalsize
    \centering
    \caption{Indicative Summary of Advanced Semantic Processing Techniques}
    \label{Summary of Advanced Semantic Processing Techniques}
        \begin{tabular}{c<{\centering}p{3cm}<{\centering}p{3.5cm}<{\centering}c<{\centering}c<{\centering}c<{\centering}p{3.7cm}<{\centering}}
        \toprule
        Reference&Sensors&Type of Raw Data&Semantics&Real Time Performance&Environments&Possible Tasks\\
        \midrule
        \cite{jang2020detection, ma2018region}&RGB camera&Colored image&Human behaviour&No&Indoor, outdoor&ER tasks, first-aid tasks\\

        \cite{simon2017hand}&RGB camera&Colored image&Gesture&Yes&Indoor, outdoor&First-aid tasks\\

        \cite{ma2016local}&RGB camera&Colored image&Human pose&No&Indoor, outdoor&First-aid tasks\\

        \cite{bosch2007image}&RGB camera&Colored image&Objects&No&Indoor, outdoor&ER tasks, heavy-duty tasks\\

        \cite{yu2020efficient}&RGB camera&Remote sensing image&Object, scene&No&Outdoor&ER tasks\\

        \cite{zhao2017fully}&RGB camera&Colored image&Material&Yes&Indoor&ER tasks, heavy-duty tasks\\

        \cite{fei2021pillarsegnet}&LiDAR&Pointcloud&Object&No&Outdoor&ER tasks\\


        \cite{christie2017radiation}&LiDAR, RGB camera, radiation detector&Pointcloud, colored image,radiation readings&Object&No&Outdoor&ER tasks\\

        \cite{zhuang2021perception}&LiDar, RGB camera&Pointcloud, colored image&Object&Yes&Outdoor&ER tasks, heavy-duty tasks\\

        \cite{zang2020far}&Thermal camera&Far-infrared image&Object&Yes&Outdoor&ER tasks\\

        \cite{pozzer2021semantic}&Thermal camera&Thermal image&Concrete condition&No&Outdoor&ER tasks, heavy-duty tasks\\

        \cite{masouleh2019development}&Thermal camera&Remote sensing thermal image&Object&Yes&Outdoor&ER tasks, First-aid tasks\\

        \cite{zhu2020intelligent}&Infrared camera&Infrared image&Object and fire status&Yes&Outdoor&Heavy-duty task\\

        \cite{shi2021semantic}&Microphone&Audio spectrogram&Scene&No&Indoor, outdoor&ER tasks\\
        
        \cite{ren2018deep}&Microphone&Audio spectrogram, scalogram&Scene&No&Indoor, outdoor&ER tasks\\
        
        \cite{okutani2012outdoor}&Microphone, RGBD camera&Audio spectrogram, colored image with depth&Location&Yes&Outdoor&ER tasks\\
        
        \cite{wu2010emotion}&Microphone&Audio spectrogram&Human mood&No&Indoor, outdoor&First-aid tasks, heavy-duty tasks\\
        
        \cite{groves2021robotic}&LiDAR, radiation detector&Pointcloud , radiation readings&Terrain, radiation location&Yes&Indoor&ER tasks, heavy-duty tasks\\
        
        \cite{hellfeld2019real}&LiDAR, RGB camera,IMU, radiation detector&Pointcloud, radiation readings&Radiation location&Yes&Indoor, outdoor&ER tasks, heavy-duty tasks\\
        
        \cite{vetter2018gamma}&RGBD camera, radiation detector&Colored image with depth, radiation readings&Radiation location&No&Indoor, outdoor&ER tasks, heavy-duty tasks\\
        
        \bottomrule
        \end{tabular}
    \end{table}
\end{landscape}

\begin{itemize}

\item Considering the features of auxiliary tasks, semantics provides limited support to auxiliary tasks.
\item Comparing with low-level semantics, high-level semantics requires more focus from researchers. There are few high-level semantic applications in the field. 
\item Some of the semantics are close to being practically used (e.g. object, distance, and scene) while others are far from obtaining semantics (e.g. SHA, human behaviour, and task workload).
\item It's clear that the semantic applications on ER tasks are more mature to be deployed in the field. For other tasks, because of their complexities, it's hard to rank their deployment possibility in the future. However, they all have big potential to involve semantic information in the sub-tasks and benefit the operation.
\item For first-aid tasks, there are techniques to obtain the basic required semantics (e.g. objects, distance, and scene). However, we still lack techniques to obtain expert knowledge (e.g. identify injuries) related semantics.

\end{itemize}
\begin{table*}
\normalsize
\caption{Current status of processing important semantics in the scope of common disaster response tasks}
\label{Gaps between task requirements and semantic understanding techniques}
\begin{threeparttable}
    \begin{tabular}{|m{1.7cm}<{\centering}|m{1cm}<{\centering}|m{1cm}<{\centering}|m{1cm}<{\centering}|m{1cm}<{\centering}|m{1cm}<{\centering}|m{1cm}<{\centering}|m{1cm}<{\centering}|m{1cm}<{\centering}|m{1cm}<{\centering}|m{1cm}<{\centering}|m{1cm}<{\centering}|}
    \hline Semantics/ Tasks &Object&Dis-tance&Status&Scene&Human Behaviour&Risk&SHA&Struc-tural Health& Task Difficulty&Traver-sability&Task Workload\\ \hline

    ER Tasks & \cellcolor[HTML]{F8CECC}T\tnote{1}\cite{wang2022yolov7}\cite{leach2021data} &\cellcolor[HTML]{F8CECC}T & \cellcolor[HTML]{EA6B66}P\tnote{2}\cite{pozzer2021semantic} &\cellcolor[HTML]{F8CECC}T\cite{zhao2017fully} & \cellcolor[HTML]{EA6B66}P\cite{ma2018region}\cite{jang2020detection} & \cellcolor[HTML]{EA6B66}P\cite{sharifi2021deep} & \cellcolor[HTML]{FF0000}N\tnote{3} & \cellcolor[HTML]{EA6B66}P\cite{pozzer2021semantic} & - \tnote{4}& \cellcolor[HTML]{F8CECC}T\cite{sevastopoulos2022survey} & \cellcolor[HTML]{FF0000}N\\ \hline

    Auxiliary Tasks & \cellcolor[HTML]{F8CECC}T & \cellcolor[HTML]{F8CECC}T & \cellcolor[HTML]{EA6B66}P & - & -& - & -& - & -  & - & -\\ \hline

    First-aid Tasks& \cellcolor[HTML]{F8CECC}T & \cellcolor[HTML]{F8CECC}T & \cellcolor[HTML]{EA6B66}P& \cellcolor[HTML]{F8CECC}T & \cellcolor[HTML]{EA6B66}P\cite{ma2016local}& \cellcolor[HTML]{EA6B66}P & \cellcolor[HTML]{FF0000}N& - & - & - & - \\ \hline


    Heavy-duty Tasks & \cellcolor[HTML]{F8CECC}T & \cellcolor[HTML]{F8CECC}T & \cellcolor[HTML]{EA6B66}P & \cellcolor[HTML]{F8CECC}T & \cellcolor[HTML]{EA6B66}P\cite{zhang2020recognition} & \cellcolor[HTML]{EA6B66}P& - & \cellcolor[HTML]{EA6B66}P & \cellcolor[HTML]{EA6B66}P\cite{sevastopoulos2022survey} & \cellcolor[HTML]{F8CECC}T & \cellcolor[HTML]{FF0000}N \\ \hline
    \end{tabular}
        \begin{tablenotes}   
            \footnotesize               
            \item[1] T: technique exist.        
            \item[2] P: partial technique achieved.
            \item[3] N: no technique.
            \item[4] -: not related. 
        \end{tablenotes}
    \end{threeparttable}
\end{table*}
The highlighted insights aim to encourage researchers to engage in grounding the advanced semantic understanding techniques on the practical disaster robotic field deployment and discover more semantic information that is useful for robots and humans.

\section{Limitations}

We note that:
\begin{itemize} 

\item Our coverage of semantics has been somewhat skewed towards image-based methods, as much of the semantics literature is based on Computer Vision (CV). This is partly because images carry more information than other carriers and semantics is a rapidly growing topic in CV, with large advances in deep learning techniques in recent years. However, we have also attempted coverage of other sensing modalities as multi-sensor fusion seems likely to play an increasingly important role to overcome the shortcomings of CV or single sensor solutions.

\item This taxonomy is based on reports of robot deployments in disaster response. Such literature is limited because disasters are themselves rare events, and robot use remains limited. As researchers, we do not have first-hand experience in disaster response. However, our team members have participated in training and field exercises with KHG, Germany's nuclear emergency robot response team and documented these experiences in \cite{manolisKHGiros2022}. We have also undertaken extensive discussions, over several years, with key personnel who deployed robots at the 2011 Fukushima nuclear disaster \cite{kawatsuma2017unitization}. We encourage more direct interaction between robotics researchers and first responders.
\end{itemize}

\section{Conclusions}
This paper aims to identify areas of potential synergy, as well as current gaps and challenges, between the AI semantic understanding community and the applied robotics disaster response domain. Our taxonomy of semantic information provides an explicit understanding of semantics' intrinsic features, and their relationships to disaster response applications. These insights suggest directions for future research in semantics understanding techniques, to robustify, adapt and apply these complex, high-level technologies to practical robot deployments in high-consequence, unstructured environments such as disaster response.

\bibliographystyle{IEEEtran}
\bibliography{cit}

\addtolength{\textheight}{-12cm}   









\end{document}